\documentclass{article}
\usepackage{spconf,amsmath,amssymb,graphicx,booktabs}
\usepackage{enumitem}
\usepackage[hidelinks]{hyperref}
\title{NOISE-ROBUST FORCED ALIGNMENT FOR SPONTANEOUS SPEECH}

\name{Laurin Wagner, Mario Zusag}
\address{nyra labs, Vienna, Austria}

\makeatletter
\def\thebibliography#1{\section{References}\list
 {[\arabic{enumi}]}{\settowidth\labelwidth{[#1]}\leftmargin\labelwidth
 \labelsep=3pt\advance\leftmargin\labelsep
 \itemsep=0pt plus 0.3pt\parsep=0pt\partopsep=0pt\topsep=1pt
 \usecounter{enumi}}
 \def\newblock{\hskip .11em plus .33em minus .07em}
 \sloppy\clubpenalty4000\widowpenalty4000
 \sfcode`\.=1000\relax}
\makeatother
\begin{document}
\ninept
\maketitle

\begin{abstract}
We show that forced aligners degrade on spontaneous, noisy real-world speech, and that fillers, laughter, and disfluencies corrupt word boundaries, impacting many downstream applications. We revisit the classical GMM-HMM aligner with upsampled WavLM features instead of MFCCs, explicit phone units for fillers and non-lexical vocal events trained on verbatim-annotated spontaneous speech, and a discriminative projection fitted on clean and noise-corrupted feature pairs to improve robustness. The pipeline is self-bootstrapped from 149\,h of curated audio without manual or external boundary targets. On conversational (Buckeye) and disfluent (FluencyBank) speech it reaches word-boundary MAEs of 20.6 and 76.9\,ms, 19\% and 17\% lower than the strongest of ten baselines. Catastrophic misalignments ($>$200\,ms) are $2.7\times$ rarer than Montreal Forced Aligners (MFA)  best model even after widening MFA's search beam, which we show removes a large pruning artifact from its conversational error; event units halve the error next to laughter, while being meaningfully more robust than existing aligners.
\end{abstract}
\begin{keywords}
forced alignment, disfluencies, conversational speech, self-supervised features, noise robustness
\end{keywords}

\section{Introduction}
\label{sec:intro}
Word- and phone-level timing are foundational signals in speech processing. Clinical speech biomarkers and fluency assessments rely on accurately timestamped transcriptions, from which pause distributions, articulation rates, and the locations of repetitions, cutoffs, and prolongations are derived~\cite{zusag2023careful,thallinger2025dementia,ratner2018fluency}. Speech corpus construction relies on precise timestamps to segment long-form audio into utterance-sized units~\cite{panayotov2015librispeech,chen2021gigaspeech,galvez2021peoples,zen2019libritts,he2024emilia,koluguri2025granary}. Beyond data preparation, accurate timing signals enable the design of training objectives: duration targets for non-autoregressive synthesis~\cite{ren2021fastspeech2}, emission-time regularization for streaming recognition~\cite{yu2021fastemit,kang2023delay}, or word-level alignment to derive speaker-invariant content representations~\cite{wagner2026pint}. Improving timing accuracy therefore strengthens the learning signal of downstream approaches that rely on such supervision, while systematic boundary errors enter it as label noise that makes training harder. On the application side, subtitling~\cite{papi2023subtitling}, automated speech editing~\cite{peng2024voicecraft,bai2022a3t}, and automatic dubbing~\cite{federico2020dubbing} are directly facilitated by accurate and robust alignment.

Existing approaches to producing such timestamps fall into several families. Classical HMM-based systems, most prominently the Montreal Forced Aligner (MFA)~\cite{mcauliffe2017montreal,mcauliffe2026mfa}, remain the de facto standard and are methodologically closest to our work: a Kaldi-based~\cite{povey2011kaldi} GMM-HMM triphone pipeline over MFCC features. Frame-classification systems score each frame with a discriminative phone classifier and recover boundaries by monotonically aligning the transcript's phone sequence to the resulting posteriors, without the HMM's transition and duration model; they require frame-level phone labels for training, either human-annotated boundaries~\cite{kelley2024maps} or the output of an HMM aligner whose errors they inherit~\cite{zhu2022charsiu}. The Qwen forced aligner~\cite{qwenfa} interleaves timestamp-slot tokens with the transcript and predicts all slots jointly in a single non-autoregressive pass over an LLM backbone, distilling and smoothing MFA-generated pseudo-labels as its timing supervision. Attention-based methods derive timestamps from decoder cross-attention patterns, popularized by Whisper~\cite{radford2023whisper} and subsequently made substantially more accurate in CrisperWhisper~\cite{wagner2024crisperwhisper} and CrisperWhisper~2.0~\cite{wagner2026policy}. These improvements arose from tokenizer changes and explicitly supervising cross-attention weights with MFA alignments during training. CTC-based approaches extract timestamps from the alignment lattices of CTC-trained ASR models~\cite{kurzinger2020ctc,bain2023whisperx,pratap2023mms,rastorgueva2023nemo}, whose peaky, blank-dominated posteriors are known to bias boundaries (visible in Sec.~\ref{ssec:main} as low tight-tolerance accuracy). Continuous integrate-and-fire models predict token firing times inside a non-autoregressive recognizer~\cite{shi2023cif}. 
Across these families, no system closes all of the following gaps jointly with sufficient quality (Table~\ref{tab:grid}). 
First, \emph{noise robustness}: MFA's MFCC front end degrades steeply under additive noise, and no aligner we are aware of explicitly learns noise-invariant alignment features. The large-scale pretraining behind CTC-based models, Whisper, CrisperWhisper~2.0, and Qwen3-FA confers partial robustness to the recognizer, but the timing mechanisms themselves (emergent cross-attention patterns; CTC posteriors, timestamps distilled from an aligner that itself fails in noise) remain unprotected. Second, \emph{explicit modeling of fillers and non-lexical vocal sound events} such as laughter or breathing: outside CrisperWhisper~2.0, these phenomena are absent from the modeled inventory, so their acoustics are easily absorbed by neighboring words or silence, corrupting boundaries. Finally, little attention is paid to the \emph{accuracy of the training annotations} themselves: systems are trained on read speech or on spontaneous speech with non-verbatim transcripts, yet verbatim annotation is what makes spontaneous speech learnable for an aligner because every untranscribed hesitation or laugh is a mislabeled stretch of audio at training time.

We close these gaps with a GMM-HMM aligner that is fully self-bootstrapped and sets a new state of the art on conversational and disfluent speech, improving over ten competitive evaluated baselines. Our contributions are:
\begin{itemize}[leftmargin=1.2em,itemsep=1pt,topsep=2pt,parsep=0pt]
\item \emph{Noise-robust feature extraction}: a WavLM-based front end with a noise-paired discriminative projection that enables accurate GMM-HMM alignment in real-world acoustic conditions.
\item \emph{Explicit modeling of spontaneous speech}: dedicated acoustic units for fillers and non-lexical vocal events, which substantially improve alignment around these phenomena.
\item \emph{Verbatim data curation}: a curated verbatim spontaneous-speech corpus, together with a quantitative evaluation focused on disfluencies and vocal events.
\end{itemize}

\begin{table}[t]
\caption{Capability grid. Lbl.-free = trained without manual or external boundary labels; $\sim$ = partial robustness inherited from augmentation or large-scale pretraining.}
\label{tab:grid}
\centering
\renewcommand{\arraystretch}{0.85}
\setlength{\tabcolsep}{1.6pt}
\begin{tabular}{lllccc}
\toprule
System & Front-end & Boundaries & Lbl.-free & Events & Noise \\
\midrule
MFA & MFCC & HMM & \checkmark & $\times$ & $\times$ \\
CTC family & w2v2 & CTC peaks & \checkmark & $\times$ & $\sim$ \\
Charsiu & w2v2 & frame & $\times$ & $\times$ & $\times$ \\
Qwen3-FA & LLM & tokens & $\times$ & $\times$ & $\sim$ \\
CrisperWhisper & Whisper & cross-attn. & $\times$ & $\times$ & $\sim$ \\
CrisperWhisper 2.0 & Whisper & cross-attn. & $\times$ & \checkmark & $\sim$ \\
\textbf{Ours} & WavLM & HMM & \checkmark & \checkmark & \checkmark \\
\bottomrule
\end{tabular}
\end{table}

\section{Training data}
\label{sec:data}
Table~\ref{tab:data} summarizes the seven training sources, totaling 149\,h, roughly half of it conversational. Conversational speech with verbatim annotations comes from the IHM splits of the ICSI~\cite{janin2003icsi} and AMI~\cite{carletta2005ami} meeting corpora, from DisfluencySpeech~\cite{wang2024disfluencyspeech}, a studio-quality single-speaker re-enactment of conversational utterances dense in disfluencies, and from part~6 of the Singapore National Speech Corpus~\cite{koh2019nsc}, which adds non-native accents. All four provide verbatim transcripts including fillers; ICSI and DisfluencySpeech further contribute annotations of vocal sounds, predominantly laughter. For read speech, Common Voice~14~\cite{ardila2020commonvoice} adds diverse recording conditions and speaker diversity, VCTK~\cite{yamagishi2019vctk} adds speaker diversity under studio conditions, and LibriSpeech~\cite{panayotov2015librispeech} contributes clean read speech. All transcripts follow the verbatim conventions of Sec.~\ref{ssec:phonemization}.

\textbf{Vocal-event overlay.} To increase the diversity and number of vocal sound events seen in training, so that the aligner can handle a larger vocal-event inventory at inference, we take clips from LibriSpeech's train-clean split in which MFA detects pauses of at least 200\,ms and insert vocal sound event clips from VocalSound~\cite{gong2022vocalsound} and Nonspeech7k~\cite{rashid2023nonspeech7k} into these pauses, adding the corresponding tags at the same transcript positions. MFA is used only to identify candidate insertion regions; its boundaries never serve as acoustic-model training targets. Since its timestamps are reliable on clean read speech (Table~\ref{tab:acc}), inserted events land in genuine silence rather than on top of speech, providing clean, well-separated examples for each event unit.

\begin{table}[t]
\caption{Training corpora. Transcripts follow a common verbatim convention; event tags are included where annotated or synthetically inserted.}
\label{tab:data}
\centering
\renewcommand{\arraystretch}{0.85}
\setlength{\tabcolsep}{2pt}
\begin{tabular}{lrrl}
\toprule
Source & Hours & Utts & Contents \\
\midrule
ICSI meetings & 39.2 & 24.7k & multi-party meetings \\
NSC part 6 & 24.5 & 4.0k & conversational, accented \\
LibriSpeech+events & 21.2 & 4.9k & audiobooks, event overlay \\
Common Voice 14 & 20.1 & 11.5k & crowd-read prompts \\
AMI meetings & 19.9 & 27.3k & multi-party meetings \\
VCTK & 15.0 & 16.2k & studio-read, many speakers \\
DisfluencySpeech & 9.2 & 4.9k & acted disfluent speech \\
\midrule
Total & 149.1 & 93.5k & \\
\bottomrule
\end{tabular}
\end{table}

\section{Method}
\label{sec:method}

\subsection{Phonemization}
\label{ssec:phonemization}
Words are converted to pronunciations with the \texttt{espeak-ng} grapheme-to-phoneme engine: its IPA output, stripped of stress and length marks, is tokenized and mapped through a hand-crafted table onto a canonical inventory of 39 English phones. This inventory is extended with dedicated units for the fillers [UH] and [UM] and twelve non-lexical vocal events (laughter, breath, cough, sigh, \dots). We adopt the verbatim transcription conventions of~\cite{wagner2026policy}, in which fillers and vocal events appear inline as bracketed tags (e.g., [UM], [laughter]); each tag maps to a one-phone lexicon entry, giving each event its own acoustics and duration statistics. The event units are supervised solely by tag occurrences in the verbatim training transcripts (Sec.~\ref{sec:data}). Where a standard aligner must learn to ignore a laugh or smear it into neighboring words, ours assigns it a unit with matching acoustics.

\subsection{Self-bootstrapped WavLM-GMM aligner}
We retain the classical Kaldi~\cite{povey2011kaldi} GMM-HMM alignment architecture of MFA-style systems and change only what the models observe. Every unit in the phoneme inventory is modeled as a three-state left-to-right HMM. Pronunciations are generated as described in Sec.~\ref{ssec:phonemization}.

\textbf{Features.} A frozen WavLM-Large~\cite{chen2022wavlm} encoder replaces the MFCC front-end. WavLM is pretrained for robustness to acoustic corruption, making it a natural candidate for extracting noise-invariant alignment features. We choose the hidden states of transformer layer 22, which have been found to correspond most closely to phonemes~\cite{chen2022wavlm} (1024-d, native 20\,ms stride). We linearly interpolate the representations to a 10\,ms grid and reduce them to 39 dimensions by a PCA fit on two million frames uniformly subsampled from our training corpus. The 39-d bottleneck mirrors the 39-dimensional MFCC front-end of the original MFA~\cite{mcauliffe2017montreal}. It is also necessary in practice: flat-start GMM training directly on the high-dimensional WavLM features was unstable in our experiments.

\textbf{Training.} The pipeline largely follows the staged monophone-to-triphone recipe of the original MFA~\cite{mcauliffe2017montreal}, with every stage supervised only by its own alignments. A monophone model is flat-started directly on the 39-d PCA features following the standard Kaldi recipe~\cite{povey2011kaldi,mcauliffe2017montreal}: the first alignment divides each utterance's frames equally among its transcript phones, and 40 EM iterations follow, realigning after every iteration with silence boosting. After obtaining monophone state alignments, we return to the WavLM representations, splice them over $\pm$1 frame, and fit a 39-dimensional linear discriminant analysis (LDA) projection~\cite{haeb1992lda} supervised by the monophone's \emph{own} per-frame state labels. In this space, a triphone model (2500 tree leaves, 6000 Gaussians) is trained for 35 EM iterations, realigning at iterations 10, 20, and 30. Unlike MFA, we apply no speaker adaptation and no maximum-likelihood linear transform (MLLT) of the features, since neither improved boundary accuracy in our ablations. After extracting the WavLM features on GPU, training completes in hours on CPU; new audio is aligned analogous to MFA.

\subsection{Noise-robust feature projection}
We train two variants that differ only in the data used to fit the LDA projection. The standard variant fits it on the clean training features. The noise-robust variant fits it on clean features together with noise-corrupted copies of the same utterances, mixed with MUSAN music~\cite{snyder2015musan} at 5--15\,dB SNR; both copies carry identical frame labels, since additive noise does not change the reference boundaries. Fisher's criterion then folds the noise variation into the within-class scatter, selecting directions that separate phones while remaining invariant to the noise.

\section{Experimental setup and metrics}
\label{sec:setup}

\textbf{Benchmarks.} We evaluate on TIMIT~\cite{garofolo1993timit} (read), Buckeye~\cite{pitt2005buckeye} (conversational), and FluencyBank Timestamped~\cite{romana2024fluencybank} (speech from people who stutter). For noise robustness, Buckeye is mixed with held-out MUSAN noise clips and synthetic Gaussian noise at signal-to-noise ratios (SNRs) of 20, 10, 5, and 0\,dB. TIMIT, Buckeye, and FluencyBank are fully held out from training.

\textbf{Protocol.} For forced aligners, predicted words correspond one-to-one to the reference transcript. For ASR-based systems, whose hypotheses can deviate from the reference, we align hypothesis and reference words with a Levenshtein alignment over canonicalized word forms and score only exactly matched words. All metrics are matched-token statistics. Non-lexical event tokens are excluded from word-level metrics for \emph{every} system for comparability. MFA baselines use CLI 3.3.9 with the latest released models \texttt{english\_mfa} 3.1.0 and \texttt{english\_us\_arpa} 3.0.0; \emph{MFA \texttt{our\_data}} is MFA's unmodified training recipe trained on our 149\,h corpus; it differs from the released models only in its training data (Sec.~\ref{sec:data}).

\textbf{Decoding beams.} MFA ships with a Viterbi beam of 10 (retry 40). We found on spontaneous speech these defaults prune the optimal path: alignments silently degrade (beam 400 with retry 4000 improves \texttt{us\_arpa} on Buckeye from 48.0 to 25.3\,ms MAE; \texttt{english\_mfa} and \texttt{our\_data} shift similarly, 48.9$\to$35.6 and 33.9$\to$24.2), and under noise most utterances fail to align at all under low beam sizes (Sec.~\ref{ssec:noise}); read speech is unaffected. Our MFA rows therefore use the wide beam unless marked. Our aligner is far more robust to this choice: in a sweep from beam 5 to 2000, its accuracy is essentially beam-independent (20.3\,ms) and converged from beam 20. The gap lies in the acoustic model: pruning discards any path falling more than the beam size below the momentary best, so a narrow beam is safe only if the emission likelihoods keep the optimal path near the top at every frame. The GMMs on our features separate phone classes sharply enough; overlapping MFCC densities let the correct path fall behind on disfluent or noisy frames, where it is pruned.

\textbf{Metrics.} For each matched word, let $\delta$ be the mean of its onset and offset timestamp errors (absolute values)  against the manual reference. We report the mean of $\delta$ (MAE). Threshold accuracies acc$_{\leq t}$, $t\in\{25,50,100,200\}$\,ms, give the share of words whose absolute offset error is at most $t$, following the convention of~\cite{rousso2024tradition, mcauliffe2026mfa}.

\section{Results}
\label{sec:results}

\subsection{Main comparison and component ladder}
\label{ssec:main}

\begin{table*}[t]
\caption{Word-boundary MAE (ms, $\downarrow$) and offset threshold accuracies (\%, $\uparrow$) on Buckeye (conversational), TIMIT (read), and FluencyBank (disfluent). All WavLM arms include the LDA projection of Sec.~\ref{sec:method}; the final row replaces it with the noise-paired LDA, all other components unchanged.}
\label{tab:acc}
\centering\small
\renewcommand{\arraystretch}{0.85}
\setlength{\tabcolsep}{3.5pt}
\begin{tabular}{l rrrrr|rrrrr|rrrrr}
\toprule
 & \multicolumn{5}{c|}{Buckeye (conversational)} & \multicolumn{5}{c|}{TIMIT (read)} & \multicolumn{5}{c}{FluencyBank (disfluent)} \\
\cmidrule(lr){2-6}\cmidrule(lr){7-11}\cmidrule(lr){12-16}
System & MAE & $\leq$25 & $\leq$50 & $\leq$100 & $\leq$200 & MAE & $\leq$25 & $\leq$50 & $\leq$100 & $\leq$200 & MAE & $\leq$25 & $\leq$50 & $\leq$100 & $\leq$200 \\
\midrule
\multicolumn{16}{@{}l}{\emph{Competing systems}} \\
Qwen3-FA-0.6B & 39.5 & 52.8 & 85.9 & 96.6 & 98.6 & 30.0 & 51.4 & 83.6 & 96.8 & \textbf{99.9} & 109.5 & 35.1 & 61.4 & 81.9 & 89.8 \\
MMS-FA & 53.0 & 57.5 & 81.3 & 92.8 & 97.8 & 37.2 & 51.5 & 82.6 & 97.5 & \textbf{99.9} & 158.8 & 35.7 & 56.8 & 74.1 & 86.4 \\
CrisperWhisper 2.0 & 38.7 & 54.2 & 78.7 & 91.7 & 97.6 & 31.2 & 58.5 & 82.6 & 92.5 & 98.8 & 93.2 & 39.8 & 62.2 & 80.3 & 90.6 \\
CTC-Segmentation & 81.0 & 41.5 & 69.5 & 89.3 & 97.4 & 49.3 & 35.3 & 64.9 & 90.9 & 99.3 & 206.6 & 22.5 & 46.0 & 68.6 & 85.0 \\
Charsiu & 70.3 & 51.8 & 68.5 & 85.1 & 94.1 & 26.6 & 68.2 & 85.8 & 95.5 & 99.6 & 198.3 & 29.9 & 49.6 & 67.2 & 79.4 \\
WhisperX & 80.9 & 35.1 & 68.1 & 94.1 & 97.8 & 77.0 & 36.9 & 68.2 & 88.0 & 95.9 & 156.8 & 26.9 & 50.9 & 77.0 & 87.7 \\
NeMo FA & 72.6 & 26.8 & 49.8 & 79.3 & 96.7 & 60.0 & 37.5 & 66.1 & 91.7 & 98.8 & 124.0 & 18.8 & 37.5 & 62.0 & 84.6 \\
\midrule
\multicolumn{16}{@{}l}{\emph{Montreal Forced Aligner family}} \\
MFA \texttt{us\_arpa} & 25.3 & 76.7 & 92.2 & 97.3 & 98.8 & \textbf{18.6} & 75.1 & 90.3 & 98.0 & \textbf{99.9} & 122.0 & 49.4 & 70.3 & 82.2 & 89.1 \\
\quad default beam & 48.0 & 75.2 & 90.4 & 95.6 & 97.2 & 20.1 & 75.2 & 90.3 & 98.0 & \textbf{99.9} & 172.3 & 49.1 & 69.8 & 81.6 & 87.8 \\
MFA \texttt{our\_data} & 24.2 & 77.1 & 92.8 & 97.7 & 98.8 & 18.8 & 77.0 & 90.8 & 97.7 & 99.8 & 136.7 & 41.8 & 62.0 & 77.6 & 86.8 \\
MFA \texttt{english\_mfa} & 35.6 & 71.3 & 87.5 & 95.2 & 97.8 & 23.2 & 69.1 & 86.3 & 96.7 & 99.6 & 162.9 & 41.8 & 63.3 & 77.0 & 85.5 \\
\midrule
\multicolumn{16}{@{}l}{\emph{Ours + component ablations}} \\
MFCC, no units & 29.3 & 74.5 & 90.3 & 96.4 & 98.6 & 22.3 & 72.6 & 88.8 & 96.3 & 99.6 & 121.0 & 44.1 & 67.8 & 83.0 & 89.9 \\
MFCC + units & 27.2 & 74.3 & 90.3 & 96.6 & 98.7 & 23.0 & 72.4 & 88.1 & 95.8 & 99.4 & 128.6 & 43.2 & 66.8 & 81.5 & 89.0 \\
WavLM (clean LDA), no units & 21.5 & 78.3 & 93.0 & 98.2 & 99.5 & 20.9 & 76.7 & 92.6 & \textbf{98.4} & 99.8 & \textbf{73.3} & 58.7 & 81.3 & 90.2 & 94.1 \\
WavLM (clean LDA) + units & \textbf{20.6} & 79.0 & \textbf{93.6} & \textbf{98.5} & \textbf{99.6} & 20.8 & \textbf{78.2} & \textbf{92.8} & \textbf{98.4} & 99.8 & 76.0 & 58.0 & 81.0 & 89.6 & 94.0 \\
\quad\textbf{with noise LDA = Ours} & \textbf{20.6} & \textbf{79.1} & 93.3 & 98.3 & \textbf{99.6} & 20.5 & 78.1 & \textbf{92.8} & \textbf{98.4} & \textbf{99.9} & 76.9 & \textbf{59.4} & \textbf{82.5} & \textbf{91.0} & \textbf{94.8} \\
\bottomrule
\end{tabular}
\end{table*}

Table~\ref{tab:acc} reports, per dataset, word-boundary MAE together with the offset threshold accuracies; rows are grouped into competing systems, the MFA family, and our system with its component ablations. We adopt the noise-robust arm as the final system: it matches the clean flagship on Buckeye while adding the robustness of Sec.~\ref{ssec:noise}.

\begin{figure}[t]
\centering
\includegraphics[width=.9\columnwidth]{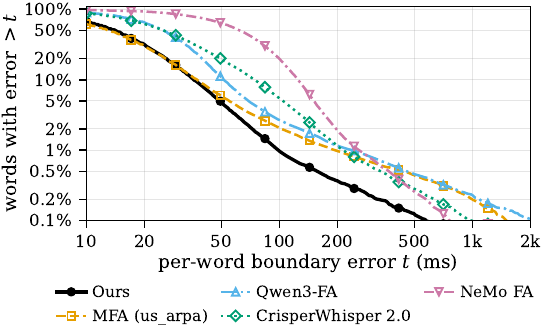}
\caption{Complementary CDF (log--log) of the per-word boundary error $\delta$ on Buckeye, all aligners at converged (wide-beam) search: each curve gives the share of words misaligned by more than $t$\,ms. Ours is the only system whose tail falls below 1\% before 100\,ms.}
\label{fig:ccdf}
\end{figure}

On read speech, we are roughly on par with the strongest classical configurations: MFA's \texttt{us\_arpa} leads by 1.9\,ms MAE (18.6 vs.\ 20.5) and produces marginally tighter boundaries at the tightest tolerances (Fig.~\ref{fig:ccdf}). On conversational and disfluent speech, baselines degrade steeply even at the wide beam: \texttt{us\_arpa} rises to 25.3\,ms on Buckeye and 122.0\,ms on FluencyBank ($6.6\times$ its read-speech error), and at MFA's package defaults to 48.0 and 172.3\,ms; MFA \texttt{our\_data} reaches 24.2 on Buckeye, indicating that data alone helps the modeling of spontaneous speech, whereas our error barely moves between read and conversational speech (20.5 vs.\ 20.6\,ms) and rises far more gently on impaired speech (76.9\,ms). This yields a 19\% margin over the best baseline on Buckeye and 17\% on FluencyBank (over CrisperWhisper~2.0, itself verbatim-trained).

Figure~\ref{fig:ccdf} shows the error profiles behind these aggregates. In the typical-word regime (left of $\sim$30\,ms) the HMM systems are nearly indistinguishable; MFA's \texttt{us\_arpa} is even marginally sharper (median 13.2 vs.\ our 14.0\,ms). From there the profiles separate: our curve runs steady below MFA's at every severity. 99\% of our words land within 97\,ms of the reference versus 196\,ms for MFA's best configuration; errors beyond 200\,ms are $2.7\times$ rarer (0.37\% vs.\ 0.98\%), beyond 1\,s $3\times$ rarer (0.07\% vs.\ 0.21\%), and averaged over each system's worst 5\% of words our error is 122\,ms against MFA's 220. At its package defaults MFA deteriorates further (99th percentile 1012\,ms, worst-5\% mean 655\,ms). The other baselines fail in characteristic ways: NeMo shows the inverse profile (imprecise everywhere, few catastrophes), Qwen3-FA shows an MFA-like mid-tail ($3\times$ our $>$200\,ms rate), and CrisperWhisper~2.0's tail still sits above ours at every threshold. This demonstrates a favorable error profile on spontaneous speech compared to existing aligners that benefits downstream use cases: typical-word precision on par with the best classical system, with much fewer severe errors.

\subsection{Alignment around disfluencies}
\label{ssec:events}

\begin{table}[t]
\caption{Buckeye MAE (ms, $\downarrow$) on regular words adjacent to a filler ($n$=2{,}656), a laugh ($n$=722), or a word cutoff ($n$=1{,}369).}
\label{tab:adj}
\centering\small
\renewcommand{\arraystretch}{0.83}
\setlength{\tabcolsep}{1.8pt}
\begin{tabular}{lrrr}
\toprule
System & filler-adj. & laugh-adj. & cutoff-adj. \\
\midrule
\multicolumn{4}{@{}l}{\emph{Competing systems}} \\
Qwen3-FA-0.6B & 56.4 & 180.4 & 70.8 \\
MMS-FA & 80.4 & 154.2 & 72.1 \\
CrisperWhisper 2.0 & 66.0 & 82.3 & 69.6 \\
CTC-Segmentation & 170.0 & 227.6 & 101.2 \\
Charsiu & 162.3 & 269.5 & 91.3 \\
WhisperX & 98.2 & 137.9 & 91.3 \\
NeMo FA & 101.1 & 105.8 & 74.8 \\
\midrule
\multicolumn{4}{@{}l}{\emph{Montreal Forced Aligner family}} \\
MFA \texttt{us\_arpa} & 40.9 & 173.8 & 38.3 \\
MFA \texttt{our\_data} & 55.2 & 120.4 & 42.5 \\
MFA \texttt{english\_mfa} & 50.9 & 230.2 & 59.9 \\
\midrule
\multicolumn{4}{@{}l}{\emph{Ours + component ablations}} \\
MFCC, no units & 47.5 & 215.6 & 42.6 \\
MFCC + units & 41.9 & 79.7 & 40.6 \\
WavLM (clean LDA), no units & 28.2 & 107.8 & 30.0 \\
WavLM (clean LDA) + units & 26.9 & \textbf{52.4} & 30.8 \\
\quad\textbf{with noise LDA = Ours} & \textbf{26.3} & 52.8 & \textbf{28.7} \\
\bottomrule
\end{tabular}
\end{table}

Table~\ref{tab:adj} isolates the regular words whose immediate neighbor is a filler, a laugh, or a word cutoff. Our aligner achieves 26--29\,ms next to fillers and cutoffs, within 10\,ms of its dataset-wide Buckeye MAE, and 53\,ms next to laughter. Even at the wide beam, every MFA configuration inflates next to a filler ($1.4$--$2.3\times$ its own dataset-wide error; \texttt{us\_arpa}: 40.9 vs.\ 25.3) and shows 120--230\,ms next to laughter ($2.3$--$4.4\times$ ours). While MFA retrained end-to-end on our data mix improves meaningfully over the released models, adding dedicated units for vocal sound events proves to be an important modeling decision, reducing laughter-adjacent error a further $2.3\times$. The component ladder separates the causes: with verbatim transcripts fillers can be represented as lexical entries, so filler units only provide a small gain on filler adjacency (47.5$\to$41.9 already in the MFCC arm; the front-end contributes the rest, $\to$28.2) but laughter has no spelling, and there the units halve the error in the WavLM arm (107.8$\to$52.4) and cut it $2.7\times$ in the MFCC arm (215.6$\to$79.7). Among baselines, only CrisperWhisper~2.0, the one other system modeling vocal events, comes near on laughter adjacency (82.3\,ms).

\subsection{Noise robustness}
\label{ssec:noise}

\begin{table}[t]
\caption{Buckeye word-boundary MAE (ms, $\downarrow$) under additive noise at SNRs of 20 to 0\,dB (column headers): held-out MUSAN noise clips and unseen Gaussian noise (training augmentation: MUSAN \emph{music} only).}
\label{tab:noise}
\centering\small
\renewcommand{\arraystretch}{0.83}
\setlength{\tabcolsep}{1.7pt}
\begin{tabular}{lrrrr|rrrr}
\toprule
 & \multicolumn{4}{c|}{MUSAN noise} & \multicolumn{4}{c}{Gaussian noise} \\
\cmidrule(lr){2-5}\cmidrule(lr){6-9}
System & 20 & 10 & 5 & 0 & 20 & 10 & 5 & 0 \\
\midrule
\multicolumn{9}{@{}l}{\emph{Competing systems}} \\
Qwen3-FA-0.6B & 41 & 54 & 79 & 154 & 43 & 85 & 154 & 339 \\
MMS-FA & 53 & 55 & 58 & 67 & 52 & 55 & 61 & 99 \\
CrisperWhisper 2.0 & 39 & 42 & 48 & 56 & 40 & 51 & 67 & 103 \\
CTC-Segmentation & 80 & 82 & 94 & 136 & 79 & 91 & 140 & 410 \\
Charsiu & 76 & 102 & 152 & 282 & 82 & 150 & 272 & 718 \\
WhisperX & 82 & 90 & 138 & 468 & 84 & 126 & 415 & 2721 \\
NeMo FA & 72 & 73 & 75 & 86 & 72 & 74 & 78 & 101 \\
\midrule
\multicolumn{9}{@{}l}{\emph{Montreal Forced Aligner family (wide beam)}} \\
MFA \texttt{us\_arpa} & 33 & 65 & 127 & 281 & 36 & 74 & 156 & 417 \\
MFA \texttt{our\_data} & 33 & 62 & 118 & 224 & 41 & 80 & 144 & 298 \\
MFA \texttt{english\_mfa} & 44 & 105 & 226 & 481 & 63 & 310 & 1037 & 2020 \\
\midrule
\multicolumn{9}{@{}l}{\emph{Ours + component ablations}} \\
MFCC, no units & 34 & 53 & 85 & 162 & 39 & 62 & 98 & 214 \\
MFCC + units & 31 & 52 & 79 & 170 & 35 & 59 & 96 & 273 \\
WavLM (clean LDA) + units & 22 & 26 & 36 & 74 & 22 & 32 & 54 & 136 \\
\quad\textbf{with noise LDA = Ours} & \textbf{21} & \textbf{22} & \textbf{26} & \textbf{50} & \textbf{20} & \textbf{25} & \textbf{36} & \textbf{96} \\
\bottomrule
\end{tabular}
\end{table}

In Table~\ref{tab:noise}, the front-end ablation isolates where robustness comes from: swapping MFCC for WavLM roughly halves the error at 0\,dB SNR under both families (MUSAN 170$\to$74\,ms, Gaussian 273$\to$136), and the noise-paired projection improves it again (50\,/\,96\,ms); since the pairing saw only MUSAN music in training, the Gaussian columns measure generalization to an unseen noise family. At default beams the MFA package fails on most utterances under heavy noise, exiting with alignment errors (\texttt{english\_mfa} retains 1.4\% of words at 0\,dB SNR under Gaussian noise); the MFA rows use the wide beam of Sec.~\ref{sec:setup}, restoring full coverage. At 0\,dB SNR under Gaussian noise, MFA \texttt{english\_mfa} misplaces words by 2.0\,s on average, \texttt{us\_arpa} by 417\,ms, MFA \texttt{our\_data} by 298\,ms, WhisperX by 2.7\,s, Qwen3-FA by 339\,ms, and ours by 96\,ms. Among CTC aligners, robustness performance tracks the stability of the acoustic model: MMS-FA and NeMo, trained on large, diverse corpora, degrade mildly (53$\to$99, 73$\to$101\,ms), while WhisperX's wav2vec2 model, trained on clean read speech, collapses. CrisperWhisper~2.0 transcribes rather than aligns: it recognizes only 29\% of words at Gaussian 0\,dB SNR, so its MAEs cover those words alone and are not directly comparable.

\section{Conclusion and Future Work}
\label{sec:conclusion}
Adjusting key design decisions in training data, phonemization, and feature extraction sets a new state of the art in spontaneous-speech alignment under noise. Multilingual verbatim transcription (CrisperWhisper~2.0) makes extending the approach beyond English feasible.

{\linespread{0.82}\selectfont\vspace{-6pt}\section{Compliance with Ethical Standards}\vspace{-3pt}
This study is a purely algorithmic contribution using publicly available benchmark datasets, for which no ethical approval was required.

\vspace{-6pt}\section{Acknowledgments}\vspace{-3pt}
No funding was received for conducting this study. The authors have no relevant financial or nonfinancial interests to disclose.

\vspace{-6pt}\bibliographystyle{IEEEbib}
\bibliography{refs}}

\end{document}